\documentclass{article}

\usepackage{amsfonts}       % blackboard math symbols

\usepackage{graphicx}       % graphics
\usepackage{graphicx,adjustbox} % for pdf, bitmapped graphics files
\usepackage{prettyref}
\usepackage{subfigure}
\usepackage{amsmath} % assumes amsmath package installed
\usepackage{amssymb}  % assumes amsmath package installed
\usepackage{color}
\usepackage{tikz-cd}
\usetikzlibrary{patterns}

\usepackage{amsmath,amsfonts,amssymb,mathrsfs}
\usepackage{epstopdf}
\newcommand{\be}{\begin{equation}}
\newcommand{\ee}{\end{equation}}
\newcommand{\ba}{\begin{array}}
\newcommand{\ea}{\end{array}}
\newcommand{\bea}{\begin{eqnarray*}}
\newcommand{\eea}{\end{eqnarray*}}

\newrefformat{sec}{Section~\ref{#1}}
\newrefformat{fig}{Figure~\ref{#1}}
\newrefformat{eq}{(\ref{#1})}

\begin{document}
\title{Relative coordinates are crucial for Ulam's ``trick to the train of thought"}

\author{Weibo Gong\thanks{The first author wishes to thank Christopher Hollot and Patrick Kelly for many discussions.}, Chirag S. Trasikar and Bradley Zylstra\\
University of Massachusetts\\
       Amherst, MA 01002, USA\\
}
\maketitle

\begin{abstract}
Spatial signal processing algorithms often use pre-given coordinate systems to label pixel positions. These processing algorithms are thus burdened by an external reference grid, making the acquisition of relative, intrinsic features difficult. This is in contrast to animal vision and cognition: animals recognize features without an external coordinate system.  We show that a coordinate system-independent algorithm for visual signal processing is not only important for animal vision, but also fundamental for concept formation. In this paper we start with a visual object deformation transfer experiment. We then formulate an algorithm that achieves deformation-invariance with relative coordinates.  The paper concludes with implications for general concept formation.
\end{abstract}

%\keywords{keyword1, keyword2, keyword3}

%\maketitle

\section{Introduction}
Recent developments of Large Language Models heated up the historical debate in AI about how a machine should generate new concepts from data corpus.  Furthermore, could the concept formation mechanism enable a thought process via recursion?  In \cite{Mitchell2023}, AI pioneer Melanie Mitchell emphasizes that being able to form new concepts and abstractions is the most important open problem in AI. As early as 1976 the recursive formation of abstract concepts in thought processes had been noted by the great scientist Stanislaw Ulam:``There must be a trick to the train of thought, a recursive formula. A group of neurons starts working automatically, sometimes without external impulse. It is a kind of iterative process with a growing pattern. It wanders about in the brain, and the way it happens must depend on the memory of similar patterns" \cite{Ulam1991}.

To address this fundamental issue it might be useful to go back to the more specific mechanism of generating concepts from visual objects in the animal brains. When a visual object deforms, the animal brain acquires the concepts of the deformations such as rotation or stretching. To achieve this, the animal brain must recognize the deformation invariants of the visual object.

The animal vision system recognizes deforming objects effortlessly, whereas machine learning algorithms have yet to match up. The fundamental difference is that the animal visual system does not use a fixed coordinate system to register the positions of the receptive fields. In contrast the computer vision algorithms use the global row and column numbers to label the image pixel positions. The information of these global coordinate numbers becomes baggage for the signal processing algorithm.  The invariant features of the visual object are mixed with such baggage which then takes layers of operations to separate if at all possible. The animal vision system must extract the invariant features early on for the quick recognition of deforming objects. The necessity of local coordinate numbers for the neighborhood relations can be met by the feature vectors that encapsulate the neighborhood content without involving a global coordinate system.  The eigenvalue vector of the Laplacian matrix formed by the content of the image patch could serve this purpose.  However, the eigenvalues are usually mixed with the associated eigenvectors in the spectral decomposition of the Laplacian matrix.  These eigenvectors are not invariant even for simple rotations. To separate the eigenvalues and eigenvectors we treat the visual signal on a small receptive field as a linear operator acting on the state of the receptive field.  This is in contrast to most spatial signal processing algorithms where the incoming visual signal is treated like an initial condition for the algorithms to act on.  We note that in the real biological scenario the visual signal does indeed act like an operator and not like an initial condition.
%However our approach was initially motivated by the state space model based %linear control system theory, as discussed later.

In this paper we propose a dynamic system model in the above spirit to guide a computer algorithm for such a separation.  The invariance acquisition is demonstrated in experiments of deformation transfer and rotated MNIST digit recognition.  The algorithm works with more general spatial signals such as those representing neuron activities, paving the way for general concept abstraction algorithms that are needed to form nodes in a symbolic AI graph.

\section{``Mental Rotation" Example}
\begin{figure}[h]
\centering
    \includegraphics[width=3.3in,height=1.0in]{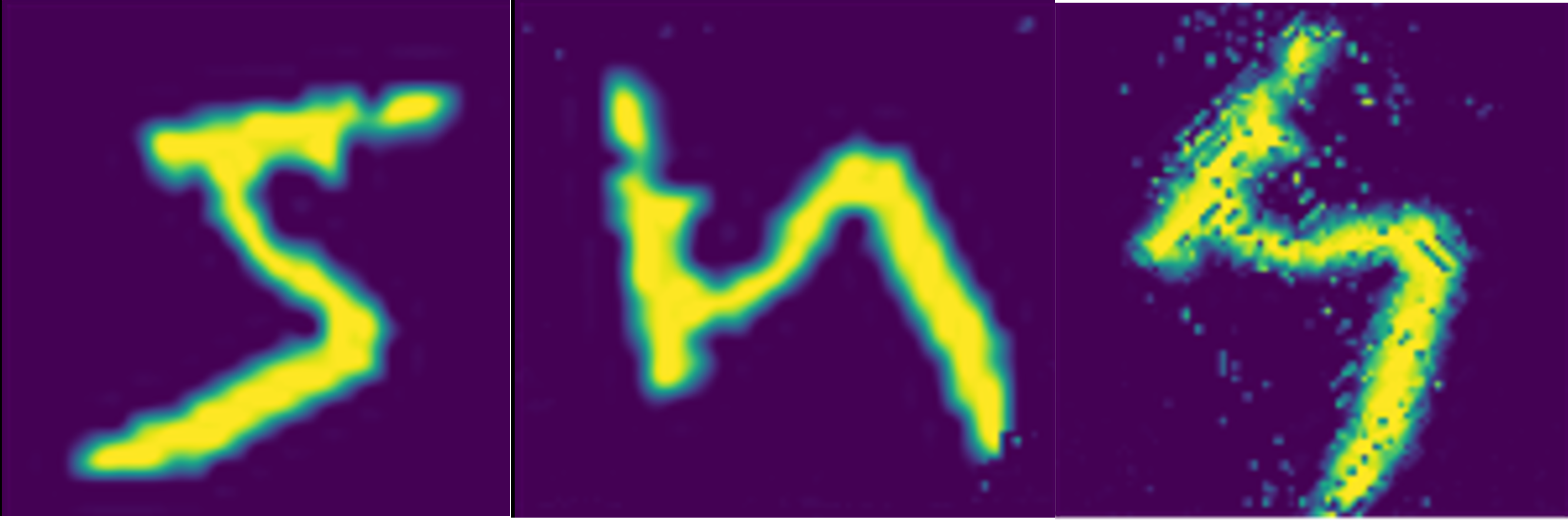}
    \caption{``Mental Rotation" of a MNIST digit}
    \label{Rotate5}
\end{figure}

\begin{figure}[h]
\centering
    \includegraphics[width=3.3in,height=0.8in]{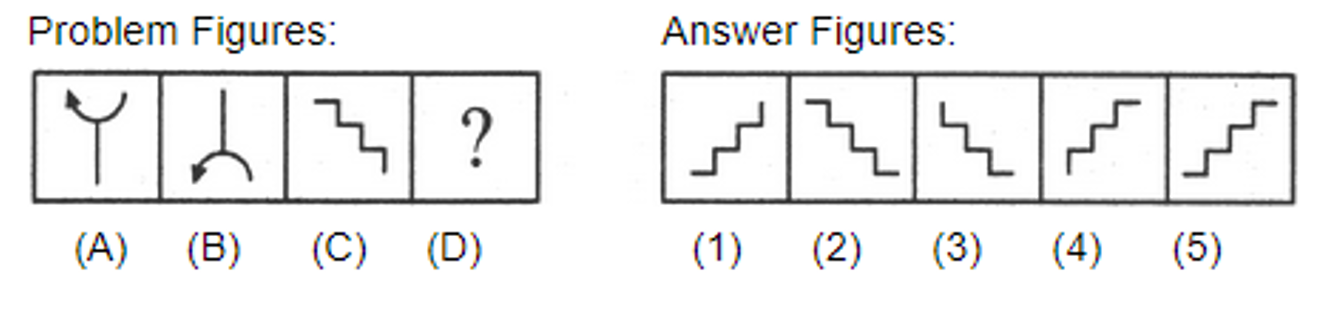}
    \caption{Non-verbal reasoning test example}
    \label{PictureIQ}
\end{figure}

\begin{figure}[h]
\centering
    \includegraphics[width=3.3in,height=1.5in]{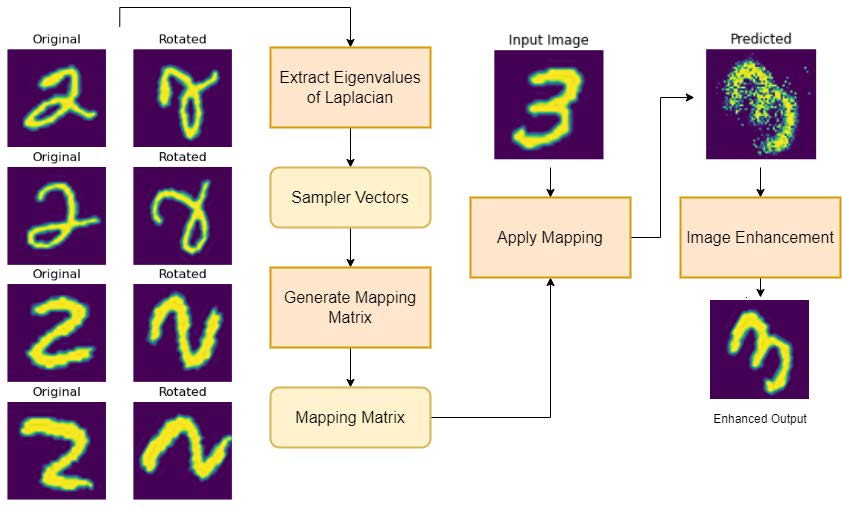}
    \caption{The 2D Deformation Transfer Pipeline. Given multiple input pairs of images and their corresponding transformed images, the algorithm applies the transformation on a new test image}
    \label{PoseTransferPipeLine}
\end{figure}

We use an MNIST digit orientation transfer example to illustrate the idea. Consider an MNIST digit 5.  We would like to make it learn rotation from some training pairs of the MNIST digit 2. Figure \ref{Rotate5} shows the results of learning to turn 90 degrees and 45 degrees from 
multiple training pairs of digit 2. Such ``mental rotation" is often used in the Raven's progressive matrix IQ test (Figure \ref{PictureIQ}.) Importantly, our algorithm does this learning without the pixel's global row and column numbers.
\begin{figure}[h]
\centering
    \includegraphics[width=3.3in,height=1.0in]{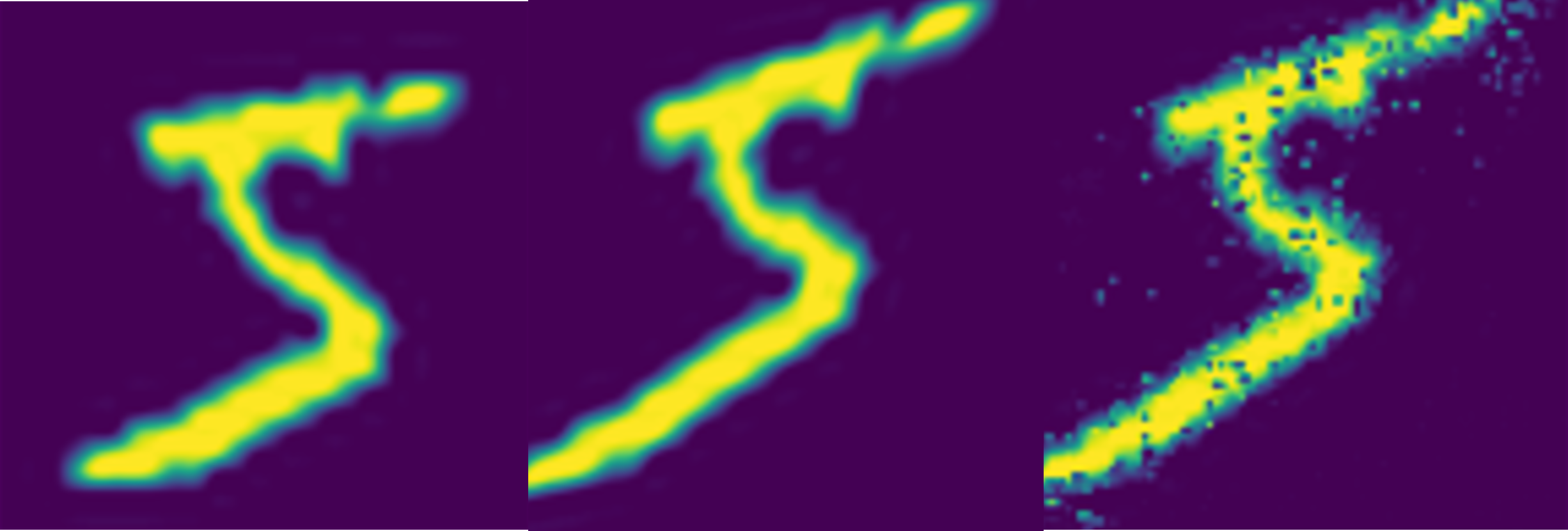}
    \caption{``Mental Affine Transformation" of a MNIST digit}
    \label{Affine5v3}
\end{figure}
A similar example of a ``mental affine transformation" is shown in Figure \ref{Affine5v3}, where the left image is the original, the center is the expected transformed digit 5 and the right is the ``mentally transformed" digit 5 as the result of learning the affine transformation from many pairs of digit 2.  Figure \ref{PoseTransferPipeLine} shows a 2D deformation transfer pipeline.
%The details of each step are explained later.
The gist of the algorithm is as follows. We first divide the digit images into small square patches of pixels. For each patch we form a Laplacian matrix. The entry is the absolute difference between any two pixels divided by their distance. We then calculate the eigenvalue vectors of the Laplacian matrices and refer them as sampler vectors.  The similarities between the sampler vectors from the original and the deformed images are used to form a mapping matrix. The mapping matrices for the individual training image pairs are averaged over. The resultant mapping matrix is used to construct the predicted image from the input image. One can see in Figure \ref{Rotate5} that the 90 degree rotation prediction is almost perfect, demonstrating the effectiveness of the mapping. The noise in the 45 degree image prediction is due to the mismatch between the rotated image patch and the upright scanning window, and can be reduced by increasing the training pairs or conventional denoising if needed.
In the Raven's progressive matrix test (Figure \ref{PictureIQ}) there is only one training pair, but it is enough for such simple images.  Here each sampler vector of Figure (A) is scanned by all the sampler vectors of Figure (B) to find a sampler vector mapping matrix. This mapping matrix compares the mapping matrices between (C) and all the answer figures to find the best fit. We note that this last step of finding the best fit among the mapping matrices is treating an ``abstract concept" representation the same way as treating the ``concrete" image signals. In this way the example serves to illustrate how the abstract thinking process is executed in the same algorithms for image processing.

%Later we will also present some rotational object recognition demos. Our %algorithm is distinct from previous effort in that we don't require a %priori knowledge of the deformation.  The key challenge we address in this %paper is deformation invariance without prior knowledge – just as Nature %does.

The polar decomposition theorem tells us that a linear transformation can be represented as the product of a rotation or reflection, and a scaling along a set of orthogonal axes. In the animal visual system the individual receptive field is small enough to approximate an object deformation in the field as a linear transformation. The scaling and reflection can be dealt with similarly as the rotations.

We note that in the above orientation transfer example the mapping matrix is the signal representation of an abstract concept that emerged from the training image pairs.  In general, a basic challenge for AI is to develop concept abstraction algorithms that can acquire the invariants of objects, whether concrete or abstract, from the instances, to form a signal representation of the concept such as the above mapping matrix. Once we know the representation of the abstract concepts we can test the similarity among them using the same algorithm for visual shape similarity testing. In the above example the mapping matrix for a 45 degree rotation can be compared with other rotation mapping matrices like in the Raven's test.  The topic of analogical reasoning for relational similarity testing among ostensibly different scenarios can be cast in this set up.

Figure \ref{PoseTransferPipeLine} illustrates the pipeline of the experiment. We first try to learn the transformation given multiple training pairs of the original images and their rotated counterparts. The concept or transformation is then stored in a mapping matrix. Once the mapping matrix is formed, it can be applied to any number of test images.

\begin{figure}[h]
\centering
    \includegraphics[width=3.3in,height=1.5in]{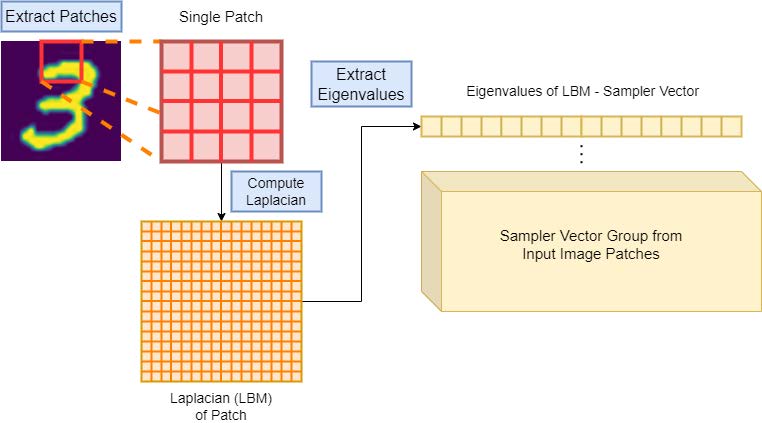}
    \caption{The Sampler Vector computation. First, we extract multiple patches from the input image. We then compute the Laplacian of each patch and then calculate the eigenvalues of each Laplacian matrix, generating an eigenvalue vector: the Sampler Vector for each patch.}
    \label{SVcomputation}
\end{figure}

We now discuss how to form the spectral feature vectors, referred as the sampler vectors,  from an image. The discussion here is tailored for the current application and is shown in Figure \ref{SVcomputation}. In short, the input image is divided into patches, simulating the multiple receptive fields on the retina. Then, the Laplacian matrix (referred in the Figure as (LBM) for Laplace-Beltrami Matrix for more general discussions) is calculated for each patch. Finally, we calculate the eigenvalues of the Laplacian matrix. In our current example the set of eigenvalues for each Laplace matrix is called the sampler vector, and we use these sampler vectors as features. From the results of our experiments it is clear that the sampler vectors do indeed encapsulate the content of each patch and this encapsulation is independent of the global coordinate numbers and orientation of the patch.
Once the sampler vectors are generated we compute the cosine similarity between the sampler vectors of the original image and the transformed image to find the matching patch pairs. The cosine similarities of the matching pairs are used to form a binary mapping matrix to predict the patch positions for constructing the transformed test image. This works thanks to the invariance of the eigenvalue vector against the image rotation. A noteworthy point here is that such invariance is achieved by the artificial order arrangement of the eigenvalue vector components, which has been made into an implicit universal convention in textbooks and numerical packages.  Nature would have to use other ways.

Generating a binary mapping matrix from the similarity matrix is not as trivial as taking the argmax along each row. This is because simply taking the argmax along the rows could map multiple patches of the transformed image to the same patch in the original image. To handle this we calculate the argmax row-wise in the order of increasing entropy. This way, the mappings for the most confident rows will be finalized first. For later rows, if the argmax is a particular column which has already been chosen for a previous row then the next best mapping is chosen.

After generating the binary mapping matrix, the mapping is applied to the patches of the new test image in the order of decreasing entropy. This way, badly mapped patches will be overlapped by better patches. Figures \ref{Rotate5} and \ref{Affine5v3} are examples. This will be further discussed in Section IV.

%XXX??XXX to continue check
\section{A dynamic system for invariants separation}
In the orientation transfer example above, the distance between two pixels in the image remains the same. In the affine transform example such distances do not change much. But the method of matching the sampler vectors can be applied more generally. For example, if the deformation is a scaling then the matching sampler vectors will live in pixel patches with correspondingly scaled sizes. This demands the sampler vectors be generated for multiple scales, a task that Nature could accomplish with the receptive field additivity. For image patches there is no such additivity and we need to calculate the sampler vectors at multiple scales.

For general cases it is useful to consider continuous time dynamic models to enlighten discrete computer algorithms. Before diving into the details of the models we highlight some useful thoughts:
\begin{itemize}
%\item {\bf to be heavily edited!!!}
\item In the brain's neural networks the visual signal elements (pixels) cannot maintain the travel timing synchronization.  The spatial signal needs to be converted to temporal signals for transmission.
\item In biological brains the similarity testing for spatial signals is hard. On the other hand the temporal signal similarity is easy to test via resonance of signal components.
\item Visual spatial signals need to be treated as operators to acquire deformation invariants.
\item To model the saccadic motions of the eyes the training images should be augmented with small changes. Invariants of an object are a voting result from the varying images of the object.
%\item ???
\end{itemize}

We now discuss these in more detail.
\subsection{Treating visual signal as an operator}
Conventional object recognition and tracking treat the image matrix as a passive signal source waiting to be processed. But in animal visual systems the light signal is actually an operator acting on the sensor panel. Since in the brain the neural circuit of surrounding-center receptive field automatically executes a Laplace action,  the image matrix should be converted to a Laplace matrix. This Laplace matrix of the image is more analogous to the system matrix $A$ of a linear system $\dot{x}(t)=Ax(t)$ than to the initial state $x(0)$, where the vector $x(t)$ is the sensor panel state. If the image matrix has a spectral decomposition $A=\sum_{k=1}^n \lambda_k\phi_k\phi^{\intercal}_k$ then the state $x(t)=\sum_{k=1}^n e^{\lambda_k t}\phi_k\phi_k^{\intercal}x(0)$. One can see that the eigenvalues $\lambda_k$ and the eigenvectors $\phi_k$ play separate roles. This has desirable implications.  For example, the eigenvalue vector $\lambda = [\lambda_1,\cdots, \lambda_N]$ of the signal matrix $A$ can be arranged as a permutation invariant vector.  When $A$ is the Laplace matrix of a small image patch, the vector $\lambda$ could be used as a local feature encapsulating the patch content and is a simple version of the ``sampler vector" of the patch. In this way an object could be represented as a set of such eigenvalue vectors.  When two such sets match well we can infer they are probably the same object.  Note again that this process is independent of the global coordinate system used to label the image pixel positions.

Note also that in real visual systems the additivity of the receptive field provides sampler vectors for multiple resolutions to enable ``cross resolution" mapping for general deformations.

\subsection{Sampler vectors for matrix images}
We first restrict to matrix image deformations that can be modeled by the pixel position changes. Consider an image matrix $I$ with pixel $(i,j)$ at the $i$th row and $j$th column. When a deformation happens we have $(i,j) \to (i',j')$, which implies that the corresponding Laplacian matrix $L$ becomes $L'$:
\begin{eqnarray*}
L'= PLP^{-1}
\end{eqnarray*}
where the permutation matrix $P$ permutes the $ij$ row and column to $i'j'$ row and column. The eigenvalues of $L$,  $\lambda_k,k=1,\cdots,N$, arranged in the conventional ascending order, are invariant to this permutation. The eigenvector permutes its components as $\phi_k=P\phi_k, k=1,\cdots,N$. We note the component sum of the eigenvectors $s_l = \sum_l \phi_k(l)$ is also invariant.  The sampler vector of $I$ is a column vector
\begin{eqnarray*}
v_k(I)= [\lambda_1,\cdots,\lambda_N, s_1,\cdots,s_N]^{\intercal}.
\end{eqnarray*}
For small image patches the eigenvector sum component of the sampler vectors can be omitted for computational convenience. These sums are similar to the phase shifts of the Fourier basis in Fourier analysis.  For small image patches these are not as important as the eigenvalues which correspond to the component magnitudes in the spectral decomposition of $L$.

The sampler vectors $v_1,\cdots,v_N$ reconstruct the Laplacian matrix $L$ up to a similarity transformation. An object consists of many small regions. Reconstructions of each small region would be enough for recognizing the object.  Now consider a linear differential equation driven by the image matrix $I$ with a uniform initial condition $\psi(x,0)$
\begin{eqnarray*}
\dot{\psi}(x,t)=L \psi(x,t).
\end{eqnarray*}
Suppose that $L$ has the spectral decomposition
 \begin{eqnarray*}
L = \sum_k \lambda_k \phi_k\phi_k^{\intercal}.
\end{eqnarray*}
We have
\begin{eqnarray*}
\psi(x,t)=\sum_k e^{\lambda_k t} \phi_k(x)\phi_k^{\intercal}(y)\psi(y,0).
\end{eqnarray*}
If we sum up the state function $\psi(x,t)$ over a region $D$ of $x$ and call it $h(t)$ we have  (assuming that $\psi(x,0)=1\ \ \forall x$)
\begin{eqnarray*}
h(t) &=& \sum_x \psi(x,t)\\
&=&\sum_k e^{\lambda_k t} \sum_x \phi_k(x)\phi_k^{\intercal}(y)\psi(y,0)\\
&=&\sum_k e^{\lambda_k t} \left(\sum_x \phi_k(x)\right)\left(
\phi_k^{\intercal}(y)\psi(y,0)\right)\\
&=& \sum_k e^{\lambda_k t} s_k^2
\end{eqnarray*}
We see that the sampler vector completely determines $h(t)$ and that the time function $h(t)$ reconstructs the system matrix $L$ up to a similarity transform from the expansion
\begin{eqnarray*}
h(t) = \sum_n \sum_x e^{Lt}\psi(x,0) = \sum_n \frac{1}{n!}\sum_x (Lt)^n.
\end{eqnarray*}
By comparing the values of $h(t)$ in the first formula where $h(t)$ is determined by the sampler vector components to the second formula where $h(t)$ is determined by the matrix $L$ and its powers, one can see that under mild conditions the sampler vector determines the entries of the matrix $L$.  We note that this conversion between the temporal and the spatial representations of the same information can be useful for parallel processing of sequential data.

The component magnitude $\lambda_k$ in the spectral decomposition of $L$ is reflected as the frequency of the ``carrier" time function $e^{\lambda_k t}$.  As mentioned, the phase information is not important for small regions and the coefficient $s_k^2$ can be taken as 1, similar to the spectrogram representation of speech signals.

\subsection{The sampler equation}
 Note that in $h(t)$ the component sum of the $k_{\rm th}$ eigenvector is ``carried" by the corresponding time function $e^{\lambda_k t}$ to provide the information of the sampler vector for the downstream receivers.  The carrier signals $e^{\lambda_i t}$ are decayed exponential functions of time and are hard to differentiate by the receivers. Second order temporal dynamics would lead to oscillatory carrier signals and the resultant sampler vector would be much more distinguishable by resonators. This motivates the following sampler equation in the continuous spatial domain:
\be
\psi_{tt}(x,t)+\gamma \psi_t(x,t) = \left(\alpha \Delta + V(x)\right)\psi(x,t),
\label{eqnpsitt}
\ee
\be
h(t)=\int_{D_s} \psi(x,t)dx.
\label{eqnh}
\ee
 Here $\psi(x,t)$ is a function of the location vector $x$ and time $t$. $V(x)$ is the image and $\gamma, \alpha$ are real constants. $D_s$ is a small region for the sampler to gather information of the spatial signal with the Laplacian $\Delta$.  This sampler converts the spatial signal $V(x)$ into an oscillatory temporal signal $h(t)$ for communication and similarity testing.  We note that $V(x)$ can be a time varying $V(x,t)$ to model the eye and image movements. The equations are based on physical mechanisms.  We treat the light signal of the image as an operator acting on the retina and the center-surround receptive field as a Laplacian. One can also imagine that the memory network is an array of resonators connected to the samplers. These resonators sift the time signals from the visual signal receptive fields according to the resonator frequencies.

 We also note that according to the Lie product formula \[e^{A+B}=\lim_{n\to\infty}(e^{A/n}e^{B/n})^n \]
 the two operators in $\left(\alpha \Delta + V(x)\right)$ can be understood as alternatively acting on the resultant state of the previous steps over very small time intervals, leading to the effect that they are acting simultaneously. In computer algorithms we approximate $\left(\alpha \Delta + V(x)\right)$ with the Laplacian matrix of the image patch.

Signal conversion via the above equations can be seen with separation of variables for the dynamic state $\psi(x,t)$:
\begin{eqnarray}
\psi(x,t)=X(x)T(t).
\end{eqnarray}
Substituting into (\ref{eqnpsitt}) one has :
\be
X(T''+\gamma T')=(\alpha \Delta +V(x))XT.
\ee
If $T''+\gamma T'=\lambda T$ and $(\alpha \Delta+V(x))X=\lambda X$ then the equation is satisfied.  The equation for the spatial variable $X(x)$ looks for the spectral decomposition of the operator $(\alpha \Delta +V(x))$:
\be
(\alpha \Delta +V(x))X=\lambda X.
\ee
The time variable $T(t)$ obeys
\be
T''+\gamma T'=\lambda T
\ee
whose solution is an exponentially decayed sinusoidal function (EDS). The spatial eigenvalue $\lambda$ and the parameter $\gamma$ jointly determine the frequency of the temporal sinusoid. The output time function $h_{D_s}(t)$ is then the sum of many terms where each term has an EDS signal with the above frequency.  The coefficients and the frequencies of these terms are the components of the sampler vector. They encapsulate the content of the visual signal in the sampling domain $D_s$. Indeed, the realization theory of linear dynamic systems confirms this. If we approximate $\alpha \Delta +V(x)$ by a large matrix and double the state space to make the equation first order in time we have a first order linear system like $\dot{x}(t)=Ax(t),y(t)=Cx(t)$. The output time function ``realizes'' the system matrix up to a similarity transform.

\iffalse
The sampler equations are meant to roughly model the energy transporting process in the early visual pathway.  The light signals form an image on the retina and the signal energy is transported through the optic nerves to the visual cortex. For each small region contributing to an optical nerve the energy of the image signal is extracted by oscillators with different frequencies. This leads to the spatial frequency decomposition of the signal in the region, since the signal energy of a particular spatial frequency $\lambda$ would be extracted by the oscillator with the time frequency $\lambda$, as described by the above equations. In short, this spatial-temporal signal conversion is a spatial frequency decomposition of the image signal. {\bf this is wrong!!!}

We add a remark here for why not using Laplace-Beltrami operator in the sampler equation. The sampler equation could be framed with a Laplace-Beltrami operator in the following way. Consider the sensor panel as a regular grid of square shaped sensors. When the light signals shine on the sensor panel the sensors perform Laplacian operation on the signals. When the light signal patterns in the visual field deform, the shapes of the sensor squares co-deform with them and keep the pattern-sensor set relations invariant. The deformation of the sensor shapes can be described by a Riemannian metric on a manifold of the light signal strength. However such models do not bring much benefits for computer implementation and we do not pursue it further here.
\fi
\subsection{Invariance of sampler vectors against image deformation}
Use of spectral invariants has a long history in computer vision. However most of the work uses a global coordinate system to set up the analysis. To illustrate the basic idea of the invariants acquisition we check some compressed celebrity images in Figure \ref{DeformationInvariants}.  We are good at recognizing celebrity faces even if the images are significantly compressed. In the figure the faces could be decomposed into Fourier transform components and we show several schematic Fourier components in the $x$ axis.  These components if stretched would recover the original face images. Comparing the purple lines between the compressed and stretched components we see that the component values at the purple lines are invariant.
%Note that these component values are the product of the normal sine function %and the Fourier magnitude at the frequency we see that the Fourier magnitudes %are invariant.
The changes happen to the corresponding spatial frequencies.  This observation of the Fourier components holds true for smooth deformations in 2D images as described below.
\begin{figure}[h]
\centering
    \includegraphics[width=3.5in,height=2in]{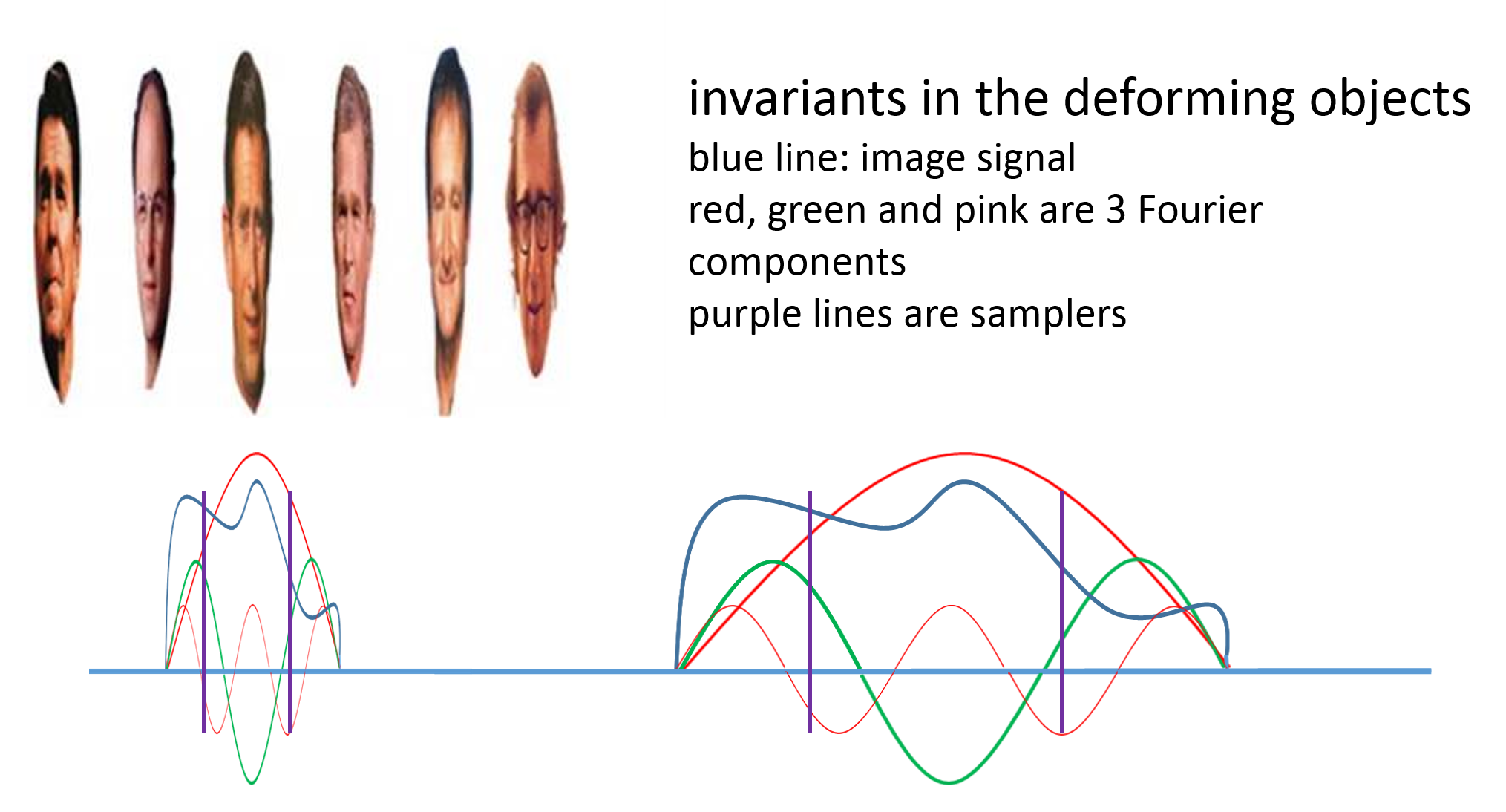}
    \caption{Invariants in simple deformations}
    \label{DeformationInvariants}
\end{figure}
Consider a square image $f(\textbf{x})$ with the coordinate grid $\textbf{x}$ and an invertible linear transform $A$. We have
\[{\cal F}f(A{\textbf x}) =  \frac{1}{|\det A|}{\cal F}(f)(A^{-\intercal}{\bf \xi})\]
where ${\cal F}$ denotes the Fourier transform operator, ${\cal F}(f)$ denotes the frequency domain function resulted from the Fourier transform acting on the spatial function $f$,  and ${\bf \xi}$ is the Fourier frequency.  The scaling factor $|\det A|$ can be removed with pixel size changes and/or frequency channel scanning. Apart from this scaling factor, the set of functions ${\cal F}f(A{\textbf x})$ is invariant against the change of $A$ in the sense that if $A \to {\tilde A}$, then one can find the values of ${\cal F}(f)(A^{-\intercal}{\bf \xi})$ at another place $({\tilde A}^{-\intercal}{\bf \xi})$. In other words, if we sample the function ${\cal F}(f)$ to form a discrete approximation vector $g(\xi_1),g(\xi_2),\cdots,g(\xi_n)$, then in the transformed image we can find a component-wise equivalent vector $g'(\xi'_1),g'(\xi'_2),\cdots,g'(\xi'_n)$ with the frequencies $\xi'_1,\xi'_2,\cdots,\xi'_n$ related to the original frequencies $\xi_1,\xi_2,\cdots, \xi_n$ according to the transform. Consequently, if one can scan the whole frequency range for similarities between the two sets of vectors, then the change of $A \to {\tilde A}$ would not affect the recognition of the original spatial function $f$.

It might be more direct to see this from
\begin{eqnarray*}
{\cal F}({\bf \xi})&=&\int f({\textbf x})e^{-{i\bf \xi}^{\intercal}{\textbf x}}dx,\\
{\cal F}(A^{-\intercal}{\bf \xi})&=&\int f(A{\textbf x})e^{-i({A^{-\intercal}{\bf \xi}})^{\intercal}{A\textbf x}}d(A{\textbf x}).
% {\cal F}(A^{-\intercal}{\bf \xi})&=&\int f(A{\textbf x})e^{-(A^{-1}{\bf %\xi})}^{-\intercal}A{\textbf x}d(A{\textbf x})
\end{eqnarray*}

The above ``global" Fourier analysis does not solve the invariance problem in real vision systems.  In a global Fourier transform any local change would affect the transformed values everywhere. In Nature or in computer algorithms one uses small receptive fields or image patches to localize the signal processing. In our setting we use the sampler vectors, which are formed from the spectral analysis of the sampler patches. These patches could be combined at different sizes to reflect the various resolutions of analysis.

\begin{figure}[h]
\centering
    \includegraphics[width=3in,height=1.3in]{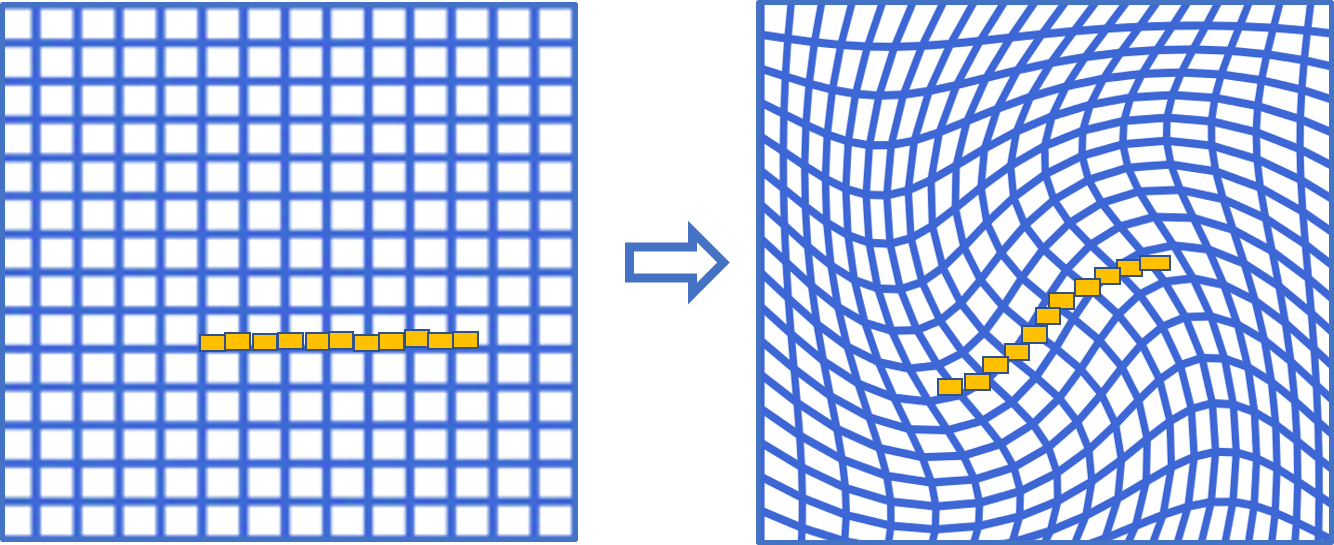}
    \caption{Laplacian sampler sets co-vary with the visual scene deformation}
    \label{LoG}
\end{figure}

In Figure \ref{LoG} we schematically illustrate a grid of samplers where a curved edge is sampled with the Laplacian samplers shown in yellow.  This group of samplers would generate the invariants since they are ``covarying" with the edge image.  Note that the position labels are based on a pre-fixed global coordinate system which does not matter to us.  Only the relative positions of the excited samplers should be in the analysis and the algorithm.

This model reflects the physical process of the visual signal processing.  The Laplacian sampler has been acknowledged to model the center-surround receptive fields. In the continuous model it is additive when covering larger visual fields. In the simplified case of the edge image in Figure \ref{LoG}, the number of Laplacian samplers would covary with the deformed edge length. In the 2D case the samplers would sample all light signals, not just the edges, but the covarying mechanism is the same.

\iffalse
\begin{figure}[h]
\centering
    \includegraphics[width=3.5in,height=2in]{DeformationInvariants}
    \caption{Invariants in simple deformations}
    \label{DeformationInvariants}
\end{figure}
\fi

We now describe the invariants acquired in the output time function of the sampler. The sampler equation for the initial visual scene $V(x)$ is:
\begin{eqnarray}
\psi_{tt}(x,t)+\gamma \psi_t(x,t) = \left(\alpha \Delta_{xx} + V(x)\right)\psi(x,t),\\
h_{D_s}(t)=\int_{D_s} \psi(x,t)dx, \hspace{1.5in}
\end{eqnarray}
where $V(x)$ is the spatial signal, $h(t)$ the temporal signal, and $D_s$ the sampling area.

The sampler equation for the deformed visual scene $U(\tilde{x})=V(x)$ (pixel value at $x$ moves to at $\tilde{x}$) is (with $\tilde{\Delta}$ representing the Laplace-Beltrami operator on a manifold \cite{rosenberg1997} at $\tilde{x}$ region):
\begin{eqnarray}
\psi_{tt}(\tilde{x},t)+\gamma \psi_t(\tilde{x},t) = \left(\alpha \tilde{\Delta}_{\tilde{x}\tilde{x}} + U(\tilde{x})\right)\psi(\tilde{x},t),\\
\tilde{h}(t)=\int_{D_{\tilde{s}}} \psi(\tilde{x},t)d\tilde{x}. \hspace{1.5in}
\end{eqnarray}
The ``Helmholtz equations" and the sampler output functions for the spetral analysis of the original and the deformed images are respectively:
\begin{eqnarray}
& & \left(\alpha \Delta_{xx} + V(x)\right)\phi_k(x)=\lambda_k\phi_k(x),\\
& & h(t)=\sum_k T_k(t) \int_{D_{s}} \phi_k(x)d(x),
\end{eqnarray}
and
\begin{eqnarray}
& & \left(\alpha \tilde{\Delta}_{\tilde{x}\tilde{x}} + U(\tilde{x})\right)\chi_l(\tilde{x})=\mu_l\chi_l(\tilde{x}),\\
& & \tilde{h}(t)=\sum_l T_l(t) \int_{D_{\tilde{s}}} \chi_l(\tilde{x})d\tilde{x}.
\end{eqnarray}

In the visual systems, the samplers are very dense with small sampler regions and therefore allow linear models. As illustrated in Figure \ref{LoG}, the set of active samplers adapts to changes in the scene. That is, the sampling is covariant with spatial deformations $U(\tilde{x})=V(x)$. The two output functions have the same ``carrier" frequencies thanks to the  covariance of the ``coordinate grid" (including the sampler domina $D_*$) and the visual signal. In other words the ``carrier" time functions $T_{*}(t)$ only differ on the ``carried" coefficients. The sampler vector consisting of the ``carrier" frequencies and the eigenvector sums are invariant. For small sampler domains we can ignore the eigenvector sums.

\section{Multiple pairs of training samples improve deformation transfer}
Consider a vision task to transfer the deformation from one object (source)  to another (target).  One can view this deformation acquisition as a primitive case of concept formation. In a real vision situation the images of the source object are always subject to small variations due to small motions of the object, the environment, and the eyes. The deformation we consider is a relatively large motion such as a 45 degree rotation. We try to understand how to form the concept of, say,  a 45 degree rotation and apply it to another object. We presented a demo case in Section II and discuss this in greater detail now.

Consider a coordinate grid ${\textbf x}$ and a real scalar function $f(x), x \in {\textbf x}$ on it.  This function $f$ is the ideal image of the source object of concern. It is subject to perturbation noises $\epsilon_i(x),i=1,\cdots,n$ and the $n$ received images are $f_i(x)= f(x)+\epsilon_i(x), i=1,\cdots,n$. Each of these source images are deformed by a smooth transformation $T: {\textbf x} \to {\textbf x'}$ into $g(x')=f(x)$.  In this discussion we are restricted to the discrete situation and the deformations are restricted to a sequence of pixel swapping that preserves the pixel pair distances during the deformation.  The rotational deformation is one such case.

If there is no noise and the $f$ values are different for different $x$ then we can find $T$ by the $f$ values. However the pixels having zero $f$ values do not contribute to the determination of $T$ since their position change does not lead to an observable change of the target object.

%In the case of constructing a deformed target object the pixels only need distinct $f$ values when their position change leads to observable %change of the target object.  For example, the pixels having zero $f$ %values do not contribute to the determination of $T$.

Now assume that we have $n$ noisy image pairs of the source and target as described before. If we simply take the average of the noisy images of the source object before the deformation as $A$ and after the deformation as $A'=TA$, and then proceed with calculations to determine $T$, we would not get good result. This is primarily because the averaging of the source object image could easily destroy the image features of the object.

Instead we can average the sampler vectors for denoising. We use data argumentation to generate multiple pairs of the source and target object and then average the sampler vectors for patches with the same the origins. The source-target object pairs should be ``covarying" during the data argumentation.  In fact, they should be treated as parts of a single object and the algorithms to extract the invariant relations is the same as if the two objects are parts of one object in a recognition task. Note that this is to mimic Nature who uses eye movement for more accurate recognition.

\section{Relevance to human cognition puzzles}
\subsection{Small number sense}
In \cite{ARC2019} Francois Chollet discusses a benchmark of tasks for machine intelligence challenge, the Abstraction and Reasoning Corpus (ARC), ``built upon an explicit set of priors designed to be as close as possible to innate human priors". One of the examples is shown below in Figure \ref{ARCexample}.  The task is to figure out the rule in the three examples shown in the left, and use the rule to solve the challenge on the right.  The solution rule is to pick the shape with the most repetitions.  This requires a machine algorithm to recognize the numbers of the same shapes in the images. Many animals are capable to sense the difference among the small numbers from one to five. It is far from clear how they do that. In fact, Albert Einstein had used small number sense as an example of concept
\begin{figure}[h]
\centering
    \includegraphics[width=3.5in,height=1.8in]{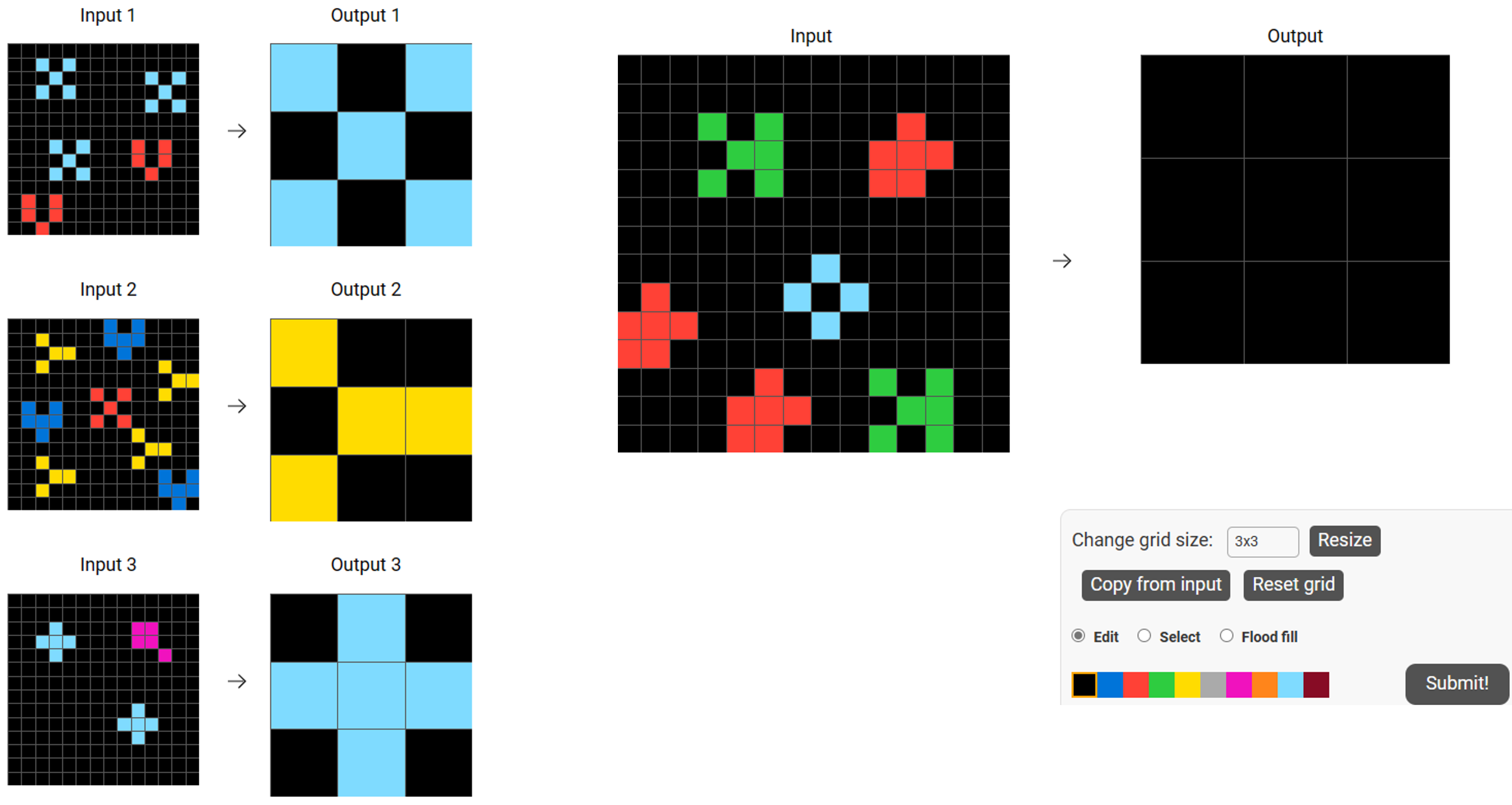}
    \caption{An ARC task that needs small number sense}
    \label{ARCexample}
\end{figure}
 emergence. ``'Three trees' is something different from 'two trees'. Again 'two trees' is different from 'two stones'. The concepts of the pure numbers 2,3,4,..., freed from the objects from which they arose, are creations of the thinking mind which describe the reality of our world.''

 \iffalse
 ``The fact that a concept in the presence of experience, even if originated from experience, has a certain logical independence is appreciated by considering extra-scientific thought. The observation of the existence of similar objects has given rise to the notion of number, but has not created it. In fact, people in some cultures have not gone any further than an understanding of only the smallest of numbers.''
\fi

Research literature has confirmed this. The Wikipedia article on parallel individuation system said the following. ``The evidence for parallel individuation system comes from a number of experiments on adults, infants and non-human animals. '' ``Parallel individuation system in animals was demonstrated in an experiment in which guppies were tested on their preference of social groups of different size, under the assumption that they have a preference for bigger size groups. In this experiment, fish successfully discriminated between numbers from 1 to 4 but after this number their performance decreased.'' 

The sampler equation (\ref{eqnpsitt}) could provide an algorithmic explanation of this.  Here the receptive field $D_s$ is the entire visual field. This large $D_s$ is available since the integration is the sum of the parts. The Laplacian matrix formed from an image with three distinct objects is a block diagonal matrix with three blocks.  Matrix theory tells us that for such a Laplacian matrix, the multiplicity of the zero eigenvalue equals to the number of blocks regardless of the block sizes. Due to the image background, the Laplacian matrix has non-zero small numbers as the off-block entries. Consequently the output time function for the image that has three objects will contain three close-to-zero frequency components that are separated from the other eigenvalues.  Therefore the visual scenes contain 2, 3 and 4 objects would form distinct neuron clusters. The similarity testing algorithm using the sampler vectors for such clusters would form the concepts of 2,3,4. As noted by Burr and Ross (2008):  ``We propose that just as we have a direct visual sense of the reddishness of half a dozen ripe cherries, so we do of their sixishness. In other words there are distinct qualia for numerosity, as there are for color, brightness, and contrast."

Further more, repeated appearances of the changes from two objects to three objects would allow a brain to form the relational matrix between 2 and 3.  The process is to fire the neural clusters representing 2 and 3 together.  Then the Spike Timing Dependent Plasticity (STDP) would form a cluster representing the concept ``2 to 3".  This cluster is also formed for 3 to 4. Similarity of the clusters ``2 to 3" and ``3 to 4" would form a new cluster representing the concept of ``plus one."  Now the brain has enough material to form the thought of applying this cluster to ``4” and call the result ``5”.  This is a ``pure mechanical" explanation of the invention of integers. This iterative process of forming abstract concepts using similarity testing of sampler vector sets generated from neuron clusters could continue further to develop arithmetics and advanced logic systems. As Leopold Kronecker put it: "God made the integers; all else is the work of man."

The importance of understanding the mechanical and algorithmic ways of generating concepts is to confirm that higher intelligence is essentially the same as lower intelligence.  As Steven Pinker \cite{Pinker2007}\cite{Pinker2010} puts it: ``We are made to be gatherers and hunters not scientists''.``The genius creates good ideas because we all create good ideas; that is what our combinatorial, adapted minds are for.''

\subsection{Commonsense psychology example}
\begin{figure}[h]
\centering
    \includegraphics[width=2.5in,height=2.0in]{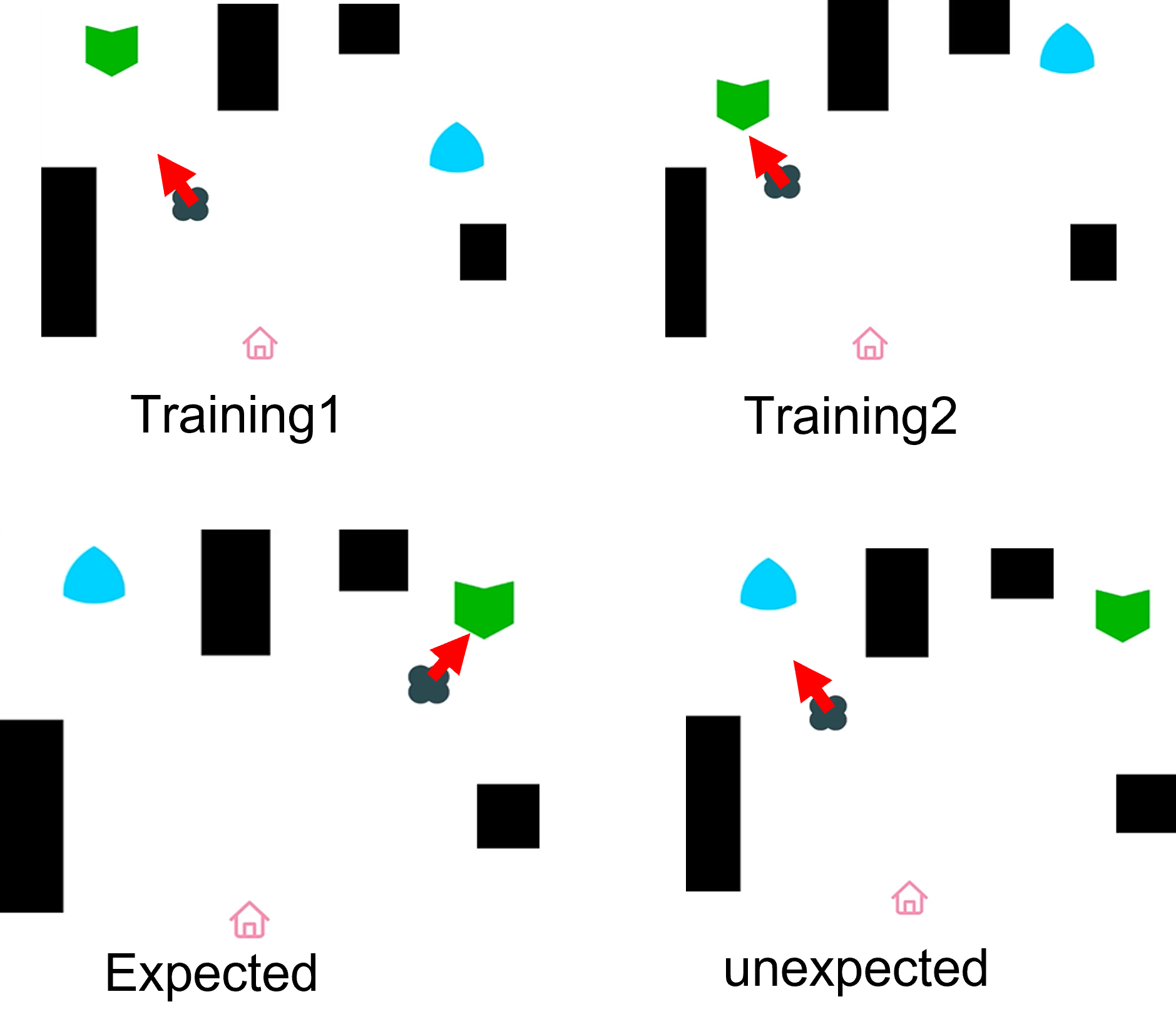}
    \caption{Commonsense psychology experiment}
    \label{Commonsense}
\end{figure}
In a recent paper \cite{Dillon2023}, the authors argued that infants' common sense psychology are foundational to human social intelligence and could thus inform better common sense AI. They experimented with infants and state-of-art AI algorithms and showed that infants are able to pick up the agent intentions while the AI algorithms failed. In Figure \ref{Commonsense} the upper row are snapshots of two out of the eight training videos.  During training the agent (gray object with red arrow added by us to indicate the motion direction in the videos) moved from the house to the target (green object). Then the infants were shown the bottom row videos.  The infants were surprised at the unexpected move of the agent going to the non-target (blue object) while not surprised when the agent is shown to move to the target (green object) despite the location swapping.

This scenario is quite similar to our mapping matrix acquisition experiments. Firstly, in our algorithm there is no global coordinate system to register the object location. Disregarding the global location of objects is a trivial consequence. Secondly, we can form the mapping matrices based on snapshots of the motion videos. The similarities and differences among these mapping matrices would then lead to reactions consistent with the experiment that the intentions are picked up during the training phase.

\section{Towards relational concept generation}
The algorithm of mapping matrix generation is intended as a prototype for general concept formation.  The recent discussions about the ChatGPT are very related.  Here we cite Steven Pinker, a leading figure in human language and psychology, and Stephen Wolfram, an expert in the computational theory of mind, to further enlighten the significance of concrete mechanisms for general concept formation.

Steven Pinker \cite{Pinker2023}: "It (ChatGPT) certainly shows how our intuitions fail when we try to imagine what statistical patterns lurk in half a trillion words of text and can be captured in 100 billion parameters. Like most people, I would not have guessed that a system that did that would be capable of, say, writing the Gettysburg Address in the style of Donald Trump. There are patterns of patterns of patterns of patterns in the data that we humans can’t fathom. It’s impressive how ChatGPT can generate plausible prose, relevant and well-structured, without any understanding of the world — without overt goals, explicitly represented facts, or the other things we might have thought were necessary to generate intelligent-sounding prose. And this appearance of competence makes its blunders all the more striking. "

Stephen Wolfram \cite{Wolfram2023}:"The success of ChatGPT is, I think, giving us evidence of a fundamental and important piece of science: it’s suggesting that we can expect there to be major new “laws of language”—and effectively “laws of thought”—out there to discover. In ChatGPT—built as it is as a neural net—those laws are at best implicit. But if we could somehow make the laws explicit, there’s the potential to do the kinds of things ChatGPT does in vastly more direct, efficient—and transparent—ways." "It’s all connected to the idea of semantic grammar—and the goal of having a generic symbolic “construction kit” for concepts, that would give us rules for what could fit together with what, and thus for the “flow” of what we might turn into human language."

\begin{figure}[h]
\centering
    \includegraphics[width=3.5in,height=1.8in]{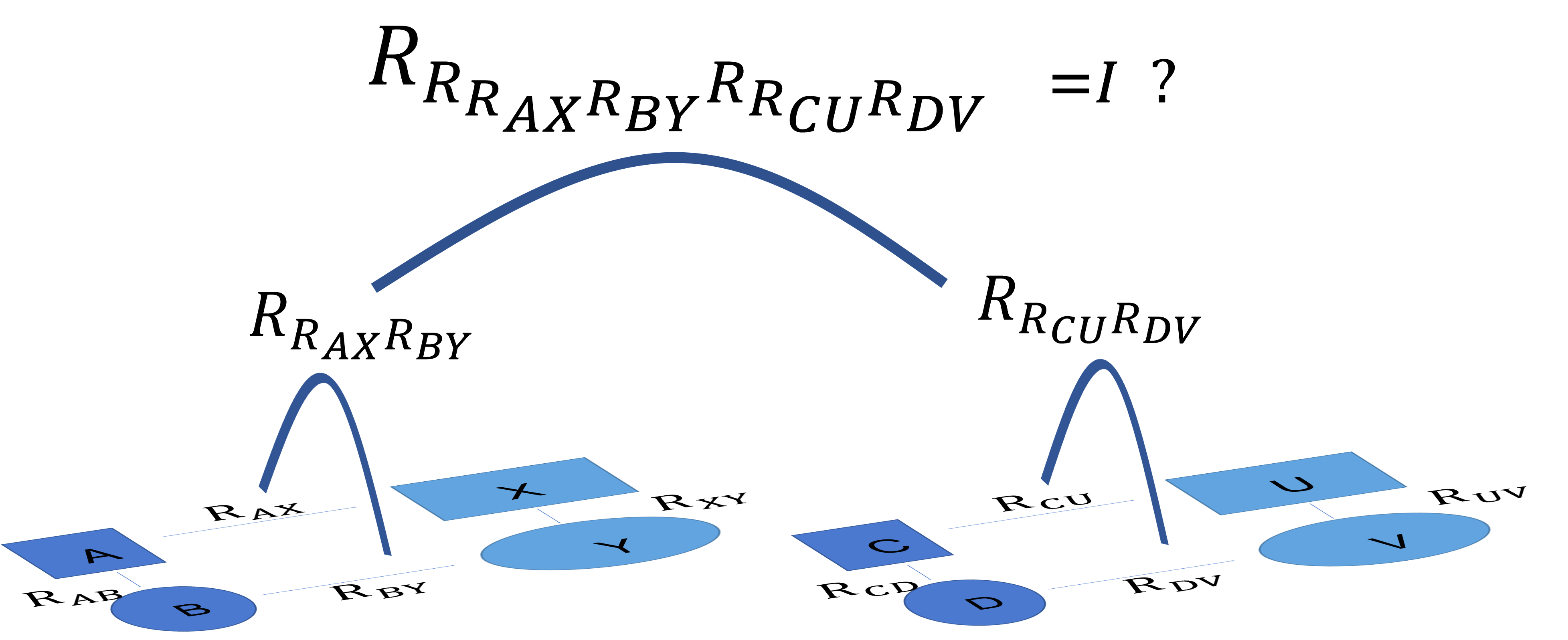}
    \caption{A Tower of Relational Concepts}
    \label{TowerofConcepts}
\end{figure}

Both experts have emphasized the importance of realizing the ``hidden patterns/rules" in human language and human thought processes. The formation of these hidden patterns have ``purely mechanical explanation" \cite{Turing1949}.  The mapping matrix in our experiments shows a prototype. The mapping matrices learned from the deformation pairs of images are themselves matrices and can be treated as images.  The algorithm for learning the ``first level" mapping matrix between images could be applied to pairs of these first level mapping matrices and generate some ``second level"  mapping matrices. Figure \ref{TowerofConcepts} illustrates this idea.
In the Figure we use capital ``R" to indicate that the mapping matrices are relational matrices. The bottom blocks marked by A, B, X, Y, C, D, U, V are images. $R_{AX}$ is the mapping matrix between the image pair A and X, and so on. The top equation is an example to check if the mapping $R_{R_{CU}R_{DV}}$ is an inverse of the mapping $R_{R_{AX}R_{BY}}$. For example, if $R_{AX}$ and $R_{BY}$ are 20 and 50 degree clockwise rotations, $R_{CU}$ and $R_{DV}$ are 20 and 50 degree anti-clockwise rotations.  Then $R_{R_{AX}R_{BY}}$ and $R_{R_{CU}R_{DV}}$ are clockwise and anti-clockwise rotation of 30 degree respectively.  The equation would hold since these two rotations would move the pixels back to their original positions. This could tell if the right side transformation compositions would give an inverse to the left side transformations.

We hope this could explain an aspect of the emergence of logic reasoning capability in animal brains.  When many such mapping matrices accumulate in the brain circuits they could form various compositions as thoughts. To this end we quote Alan Turing on analogy in an interview on October 27, 1949 \cite{Turing1949}, since the ``purely mechanical explanation" of analogical thinking he mentioned could be achieved with the similarity testing of the mapping matrices:

``I think you could make a machine spot an analogy, in fact it's quite a good instance of how machine could be made to do some of those things that one usually regards as essentially a human monopoly. Suppose that someone was trying to explain the double negative to me, for instance, that when something isn't not green it must be green, and he couldn't quite get it across. He might say ‘well, it's like crossing the road. You cross it, and then you cross it again, and you are back where you started.’  This remark might just clinch it. This is one of the things one would like to work with machines,  and I think it would be likely to happen with them. I imagine that the way analogy works in our brains is something like this. When two or more sets of ideas have the same pattern of logical connections,  the brain may very likely economize parts of by using some of them twice over,  to remember the logical connections both in the one case and in the other. One must suppose that some part of my brain was used twice over in this way, once for the idea of double negation and the once for crossing the road, there and back.  I'm really supposed to know about both these things but can't get what it is the man is driving at, so long as if he is talking about all those dreary nots and not-nots. Somehow it doesn't get through to the right part of the brain but as soon as he says his piece about crossing the road it gets through to the right part of the brain. But as soon as he says his piece about crossing the road it gets through to the right part, but by a different route.  If there is some such a purely mechanical explanation of how this argument by analogy goes on in the brain, one could make a digital computer to do the same."

\section{Conclusion}

We emphasize the following points:
\begin{itemize}
\item Visual signal processing should address pixels with the neighborhood content;
\item Content addressable processing enables recursive concept formation for general neuronal signals;
\item Relational matrices formed via recursion provide rich material for the thought process. Many of such matrices represent neural signals that are not consciously perceivable;
\item The repeated relational signals are memorized in the network weights via Hebb's rule or backpropagation and will exert influences on the future outputs.
\end{itemize}

\bibliographystyle{plain}
\bibliography{refV8}

\end{document}